\documentclass[letterpaper]{article}
\usepackage{microtype}
\usepackage{graphicx}
\usepackage{subfigure}

\usepackage{hyperref}

\usepackage{graphicx}
\usepackage[nolinenums]{custom_arxiv}
\usepackage{thm-restate}
\usepackage{booktabs}

\usepackage{amsmath}
\usepackage{amssymb}
\usepackage{amsthm}
\usepackage{float}
\usepackage{tikz}
\usepackage{graphicx}
\usepackage{enumitem}

\DeclareMathOperator{\co}{co}
\DeclareMathOperator*{\argmin}{arg\,min}

\newtheorem{definition}{Definition}
\newtheorem{lemma}{Lemma}
\newtheorem{corollary}{Corollary}

\usetikzlibrary{fit,arrows,calc,positioning,shapes.geometric}
\usetikzlibrary{arrows,arrows.meta}

\DeclareRobustCommand{\treeplexproduct}{%
  \tikz[]{%
    \draw (0,0) circle[radius=1.1mm];%
    \draw (-.55mm, -.55mm) -- (.55mm, .55mm);%
    \draw (-.55mm, .55mm) -- (.55mm, -.55mm);%
  }%
}
\DeclareRobustCommand{\computeloss}{%
  \tikz[thick,scale=.25]{%
    \draw[] (0, 0) -- (1, 0) -- (1, 1) -- (0, 1) -- (0, 0) -- (1.0, 0);%
    \draw[] (0, 0) -- (1, 1);%
  }%
}
\DeclareRobustCommand{\computerecc}{%
  \tikz[thick,scale=.25]{%
    \draw[] (0, 0) circle[radius=.5];%
    \draw[] (-.35,-.35) -- (.35,.35);%
    \draw[] (-.35,.35) -- (.35,-.35);%
  }%
}
\DeclareRobustCommand{\computesum}{%
  \tikz[thick,scale=.28]{%
    \draw[] (0, 0) circle[radius=.5];%
    \draw[] (-.3,0) -- (.3,0);%
    \draw[] (0,.3) -- (0,-.3);%
  }%
}

\makeatletter
\g@addto@macro \normalsize {%
 \addtolength\abovedisplayskip{-1pt}%
 \addtolength\belowdisplayskip{-1pt}%
}
\makeatother

\newcommand{\cF}{\mathcal{F}}

\DeclareMathSizes{10}{9}{7}{5}

\usepackage{etoolbox}
\ifcsmacro{assumption}{}{
	
}
\ifcsmacro{corollary}{}{
	
}
\ifcsmacro{definition}{}{
	\newtheorem{definition}{Definition}[]
}
\ifcsmacro{example}{}{
	
}
\ifcsmacro{fact}{}{
	
}
\ifcsmacro{informal_theorem}{}{
	
}
\ifcsmacro{lemma}{}{
	
}
\ifcsmacro{proposition}{}{
	
}
\ifcsmacro{remark}{}{
	
}
\ifcsmacro{theorem}{}{
	\newtheorem{theorem}{Theorem}[]
}

\usepackage{mathtools}
\mathtoolsset{centercolon}
\newcommand{\defeq}{\mathrel{:\mkern-0.25mu=}}

\newcommand{\bbR}{\ensuremath{\mathbb{R}}}

\newcommand{\R}{\ensuremath{\mathbb{R}}}

\newcommand{\cX}{\ensuremath{\mathcal{X}}}
\newcommand{\cY}{\ensuremath{\mathcal{Y}}}

\newcommand{\cfrp}{CFR$^+$}
\newcommand{\rmp}{RM$^+$}

\newcommand{\be}{\begin{eqnarray}}
\newcommand{\ee}[1]{\label{#1}\end{eqnarray}}

	\newcommand{\ese}{\end{eqnarray*}}
	\newcommand{\bse}{\begin{eqnarray*}}
	\def\beq{\begin{equation}}
	\def\eeq{\end{equation}}

	\def\fnote#1{\footnote}

	\def\*{{{\LARGE\bf $^*$}}}

	\def\R{{\mathbb{R}}}

	\def\cF{{\cal F}}

	\def\cL{{\cal L}}

	\def\cX{{\cal X}}
	\def\cY{{\cal Y}}

	\def\argmin{\mathop{\rm argmin}}

	\def\co{\mathop{{\rm co}}}

\title{Regret Circuits: Composability of Regret Minimizers}
\author{
  Gabriele Farina\\
  Computer Science Department\\
  Carnegie Mellon University\\
  Pittsburgh, PA 15213 \\
  \texttt{gfarina@cs.cmu.edu}\\
   \And
  Christian Kroer\\
  IEOR Department\\
  Columbia University\\
  New York NY 10027\\
  \texttt{christian.kroer@columbia.edu}\\
  \And
  Tuomas Sandholm\\
  Computer Science Department\\
  Carnegie Mellon University\\
  Pittsburgh, PA 15213 \\
  \texttt{sandholm@cs.cmu.edu}\\
}

\begin{document}
\twocolumn[\maketitle]
\begin{abstract}
  Regret minimization is a powerful tool for solving large-scale problems; it
  was recently used in breakthrough results for large-scale extensive-form game
  solving. This was achieved by composing simplex regret minimizers into an
  overall regret-minimization framework for extensive-form game strategy spaces.
  In this paper we study the general composability of regret minimizers. We
  derive a calculus for constructing regret minimizers for composite convex sets
  that are obtained from convexity-preserving operations on simpler convex
  sets. We show that local regret minimizers for the simpler sets
  can be combined with additional regret minimizers into an aggregate regret
  minimizer for the composite set. As one application, we show that
  the CFR framework can be constructed easily from our framework. We also show ways to include curtailing (constraining) operations into our framework. For one, they enables the construction of CFR generalization for extensive-form games with general convex strategy constraints that can cut across decision points.
\end{abstract}

\section{Introduction}
\emph{Counterfactual regret minimization (CFR)}~\citep{Zinkevich07:Regret}, and
its newer variants~\citep{Lanctot09:Monte,Brown15:Regret,Tammelin15:Solving,Brown17:Dynamic,Brown17:Reduced,Brown19:Solving}, have been a
central component in several recent milestones in solving imperfect-information
\emph{extensive-form games (EFGs)}. \citet{Bowling15:Heads} used {\cfrp} to
near-optimally solve heads-up limit Texas hold'em. \citet{Brown17:Superhuman}
used CFR variants, along with other
scalability techniques such as real-time endgame solving~\citep{Ganzfried15:Endgame,Burch14:Solving,Moravcik16:Refining,Brown17:Safe} and automated action abstraction~\cite{Brown14:Regret}, to create \emph{Libratus}, an AI that beat top human specialist professionals at
the larger game of heads-up no-limit Texas hold'em. \citet{Moravvcik17:DeepStack} also used CFR variants and endgame solving to beat professional human players at that game.

CFR and its newer variants are usually presented as algorithms for finding an approximate Nash equilibrium in zero-sum EFGs. However, an alternative view is that it is a framework for constructing regret minimizers for the types of action spaces encountered in EFGs, as well as single-agent sequential decision making problems with similarly-structured actions spaces. Viewed from a convex optimization perspective, the class of convex sets to which they apply are sometimes referred to as \emph{treeplexes}~\citep{Hoda10:Smoothing,Kroer15:Faster,Kroer18:Faster}. In this view, those algorithms specify how a set of regret minimization algorithms for simplexes and linear loss functions can be composed to form a regret minimizer for a treeplex. \citet{Farina19:Online} take this view further, describing how regret-minimization algorithms can be composed to form regret minimizers for a generalization of treeplexes that allows convex sets and convex losses. This decomposition into individual optimization problems can be beneficial because it enables the use of 1) different algorithms for different parts of the search space, 2) specialized techniques for different parts of the problem, such as warm starting~\citep{Brown14:Regret,Brown15:Simultaneous,Brown16:Strategy} and pruning~\citep{Lanctot09:Monte,Brown15:Regret,Brown17:Dynamic,Brown17:Reduced}, and 3) approximation of some parts of the space.

In this paper we introduce a general methodology for composing regret minimizers. We derive a set of rules for how regret minimizers can be constructed for composite convex sets via a \emph{calculus} of regret minimization: given regret minimizers for convex sets $\cX,\cY$ we show how to compose these regret minimizers for various convexity-preserving operations performed on the sets (e.g., intersection, convex hull, Cartesian product), in order to arrive at a regret minimizer for the resulting composite set.
\footnote{This approach has parallels with the calculus of convex sets and functions found in books such as \citet{Boyd04:Convex}. It likewise is reminiscent of \emph{disciplined convex programming}~\citep{Grant06:Disciplined}, which emphasizes the solving of convex programs via composition of simple convex functions and sets. This approach has been highly successful in the CVX software package for convex programming~\citep{Grant08:Cvx}.}

Our approach treats the regret minimizers for individual convex sets as black boxes, and builds a regret minimizer for the resulting composite set by combining the outputs of the individual regret minimizers. This is important because it allows freedom in choosing the best regret minimizer for each individual set (from either a practical or theoretical perspective). For example, in practice the \emph{regret matching}~\citep{Hart00:Simple} and \emph{regret matching}$^+$ ({\rmp})~\citep{Tammelin15:Solving} regret minimizers are known to perform better than theoretically-superior regret minimizers such as \emph{Hedge}~\citep{Brown17:Dynamic}, while Hedge may give better theoretical results when trying to prove the convergence rate of a construction through our calculus.

One way to conceptually view our construction is as \emph{regret circuits}: in order to construct a regret minimizer for some convex set $\cX$ that consists of convexity-preserving operations on (say) two sets $\cX_1,\cX_2$, we construct a regret circuit consisting of regret minimizers for $\cX_1$ and $\cX_2$, along with a sequence of operations that aggregate the results of those circuits in order to form an overall circuit for $\cX$. We use this view extensively in the paper; we show the regret-circuit representation of every operation that we develop.

As an application, we show that the correctness and convergence rate of the CFR algorithm can be proven easily through our calculus. We also show that the recent \emph{Constrained CFR} algorithm~\citep{Davis19:Solving} can be constructed via our framework.
Our framework enables the construction of two algorithms for that problem.
The first is based on Lagrangian relaxation, and only guarantees approximate feasibility of the output strategies.
The second is based on projection and guarantees exact feasibility, for the first time in any algorithm that decomposes overall regret into local regrets at decision points.

\section{Regret Minimization}
We will prove our results in the online learning framework called \emph{online convex optimization}~\citep{Zinkevich03:Online} (OCO).
%In this section, we will briefly touch on the important ideas used in this paper.
In OCO, a decision maker repeatedly interacts with an unknown environment by making a sequence of decisions $x^1$, $x^2, \dots$ from a convex and compact set $\cX \subseteq \mathbb{R}^n$. After each decision $x^t$, the decision maker faces a convex \emph{loss function} $\ell^t(x^t)$, which is unknown to the decision maker until after the decision is made. So, we are constructing a device that supports two operations:
(i) it provides the next decision $x^{t+1}\!\in\!\cX$ and
(ii) it receives/observes the convex loss function $\ell^t$ used to ``evaluate'' decision $x^t$.
The decision making is \emph{online} in the sense that the next decision, $x^{t+1}$, is based only on the previous decisions $x^1, \dots, x^t$ and corresponding observed loss functions $\ell^1, \dots, \ell^t$.

The quality of the device is measured by its \emph{cumulative regret}, which is the difference between the loss cumulated by the sequence of decisions $x^1, \dots, x^T$ and the loss that would have been cumulated by playing the best-in-hindsight time-independent decision $\hat x$. Formally, the cumulative regret up to time $T$ is
\begin{equation}\label{eq:regret defn}
  R_{(\cX,\cF)}^T \defeq \sum_{t=1}^T \ell^t(x^t) - \min_{\hat x \in \cX} \left\{\sum_{t=1}^T \ell^t(\hat x)\right\}\!.
\end{equation}%
Above we introduce the new notation of a subscript $(\cX, \cF)$ to be explicit about the domain of the decisions $\{x^t\}$ and the domain of the loss functions $\{\ell^t\}$, respectively. This turns out to be important because we will study composability of devices  with different domains.

The device is called a \emph{regret minimizer} if it satisfies the desirable property of
\emph{Hannan consistency}: the average regret approaches zero, that is, $R_{(\cX,\cF)}^T$ grows \emph{sublinearly} in $T$. Formally, in our notation, we have the following definition.
%The above discussion can be formalized as the following definition, where again we are careful about the domains of the decisions and loss functions.
\begin{definition}[$(\cX, \cF)$-regret minimizer]\label{def:XF regret minimizer}
  Let $\cX$ be a convex and compact set, and let $\cF$ be a convex cone in the space of bounded convex functions on $\cX$, and such that $\cF$ contains the space $\mathcal{L}$ of linear functions.
  An $(\cX, \cF)$-\emph{regret minimizer} is a function that selects the next decision $x^{t+1} \in \cX$ given the history of decisions $x^1, \dots, x^{t}$ and observed loss functions $\ell^1, \dots, \ell^t \in \cF$, so that the cumulative regret $R_{(\cX, \cF)}^T = o(T)$.
\end{definition}

\subsection{Universality of Linear Loss Functions}\label{sec:universality linear losses}
Regret minimizers for linear loss functions are in a sense \emph{universal}: one can construct a regret minimizer for convex loss functions from any regret minimizer for linear loss functions (e.g.,~\citet{Mcmahan11:Follow}).
The crucial insight is that
the regret that we are trying to minimize, $R_{(\cX, \cF)}^T$, is bounded by the regret of a $(\cX,\cL)$-regret minimizer that, at each
time $t$, observes as its loss function a tangent plane of $\ell^t$ at the most recent decision $x^t$.
Thus we can minimize $R_{(\cX, \cF)}^T$ by minimizing $R^T_{(\cX,\cL)}$.

Formally, let $\partial \ell^t(x^t)$ be any subgradient of $\ell^t$ at $x^t$. By convexity of $\ell^t$,
\[
  \ell^t(\hat x) \ge \ell^t(x^t) + \langle \partial \ell^t(x^t),\, \hat x - x^t\rangle \quad\forall\,\hat x\in\cX,
\]
and, substituting into~\eqref{eq:regret defn}, we obtain% in the notation of our Definition~\ref{def:XF regret minimizer}:
\begin{align}
    R_{(\cX,\cF)}^T
       %&\le \sum_{t=1}^T\! \ell^t(x^t) \!-\! \min_{{\hat x}\in\mathcal{X}} \left\{\sum_{t=1}^T \ell^t(x^t) + \langle \partial \ell^t(x^t),\, \hat x - x^t\rangle\!\right\}\nonumber\\[-1mm]
       &\le \sum_{t=1}^T \langle \partial \ell^t(x^t),\, x^t \rangle - \min_{{\hat x} \in \mathcal{X}} \left\{\sum_{t=1}^T \langle \partial \ell^t(x^t),\, \hat x\rangle\!\right\}\!,\label{eq:linearization bound}
\end{align}%
where the right hand side is $R^T_{(\cX,\cL)}$, the regret cumulated by a device that observes the \emph{linear} loss functions $\langle \partial \ell^t(x^t),\cdot\,\rangle$.\footnote{A downside of this approach is that~\eqref{eq:linearization bound} is an inequality, not an equality. When we use a linearization of the loss function at each decision point, we introduce error. This can cause regret to be minimized more slowly than somehow working on the nonlinear loss functions directly. Nevertheless, we obtain a regret minimizer for the original problem.
%\todo{Note on boundedness}
}

\subsection{Connection to Convex-Concave Saddle-Point Problems and Game Theory}
In this subsection we review how regret minimization can be used to compute solutions to \emph{regularized bilinear saddle-point problems}, that is solutions to problems of the form
\begin{align}
  \min_{x\in \cX} \max_{y \in \cY} \big\{ x^{\!\top}\! A y + d_1(x) - d_2(y) \big\}, \label{eq:convex concave EFG}
\end{align}%
where $\cX,\cY$ are closed convex sets, and $d_1,d_2$ are convex functions.
This general formulation allows us to capture, among other settings, several game-theoretical applications such as computing Nash equilibria in two-player zero-sum games. In that setting, $d_1$ and $d_2$ are the constant zero functions, $\cX$ and $\cY$ are convex polytopes whose description is provided by the \emph{sequence-form constraints}, and $A$ is a real payoff matrix~\citep{Stengel96:Efficient}.

In order to use regret minimization to solve problems of the form~\eqref{eq:convex concave EFG}, we consider the loss functions
% $\ell^t_\cX:\cX\to\mathbb{R}$ and
% $\ell^t_\cY:\cY\to\mathbb{R}$,
% defined as
%
\begin{align*}
  \ell^t_\cX : \cX \ni x &\mapsto  (-Ay^{t})^{\!\top} x + d_1(x), \\
  \ell^t_\cY : \cY \ni y &\mapsto  (A^\top x^{t})^{\!\top} y + d_2(y).
\end{align*}
%
% With this choice of loss function, the induced
% regret-minimizing dynamics for the two players lead to a convex-concave
% saddle-point problem.
% Specifically, assume the two players play the game $T$
% times, accumulating regret after each iteration as in Figure~\ref{fig:no alternation}.
% %
% \begin{figure}[htb]
%   \centering\includegraphics[scale=.85]{figs/no_alternation.pdf}
%   \caption{The flow of strategies and loss functions in regret minimization for games.}
%   \label{fig:no alternation}
% \end{figure}
%
% A folk theorem explains the tight connection between
% low-regret strategies for the choice of losses above and approximate Nash equilibria. We will need a more
% general variant of that theorem generalized to \eqref{eq:convex concave EFG}.
The error metric that we use is the  \emph{saddle-point
  residual} (or \emph{gap}) $\xi$ of $(\bar x, \bar y)$, defined as
\begin{align*}
  \max_{\hat{y}\in\cY} \{ d_1\!(\bar x) \!-\! d_2(\hat{y}) \!+\! \langle \bar x, A \hat y\rangle\! \} \!-\! \min_{\hat x\in\cX} \{ d_1\!(\hat x) \!-\! d_2(\bar y) \!+\! \langle \hat x, A \bar y \rangle\! \}.
\end{align*}%
The following folk theorem shows that the average of a sequence of regret-minimizing strategies for the choice of losses above leads to a bounded saddle-point residual (see, for example, \citet{Farina19:Online} for a proof).
\begin{theorem}\label{thm:folk theorem}
  If the average regret accumulated on $\cX$ and $\cY$ by the two sets of strategies $\{x_t\}_{t=1}^T$ and $\{y_t\}_{t=1}^T$ is $\epsilon_1$ and $\epsilon_2$, respectively, then the strategy profile $(\bar x, \bar y)$ where
  $
    \bar x = \frac{1}{T}\sum_{t=1}^T x^t,\ \bar y = \frac{1}{T} \sum_{t=1}^T y^t
  $
  has a saddle-point residual bounded above by $\epsilon_1 + \epsilon_2$.
\end{theorem}
%%
% The above averaging works because
% that spaces $\cX$ and $\cY$ are also convex. After averaging we can easily compute $\bar{x}$ in
% linear time.
%
When $d_1 \equiv d_2 \equiv 0$ and $\cX,\cY$ are the players' sequence-form strategy spaces, Theorem~\ref{thm:folk theorem} asserts that the average strategy profile produced by the regret minimizers is an $(\epsilon_1 + \epsilon_2)$-Nash equilibrium.

Different choices of the regularizing functions $d_1$ and $d_2$ can be used to solve for strategies in other game-theoretic applications as well, such as computing a normal-form quantal-response equilibrium~\citep{Ling18:What,Farina19:Online}, and several types of opponent exploitation. \citet{Farina19:Online} study opponent exploitation where the goal is to compute a best response, subject to a penalty for moving away from a precomputed Nash equilibrium strategy; this is captured by having $d_1$ or $d_2$ include a penalty term that penalizes distance from the Nash equilibrium strategy. \citet{Farina17:Regret} and~\citet{Kroer17:Smoothing} study constraints on individual decision points, and \citet{Davis19:Solving} study additional constraints on the overall EFG polytopes $\cX,\cY$. Regret minimization in those settings requires regret minimizers that can operate on more general domains $\cX,\cY$ than the sequence form. In this paper we show how one can construct regret minimizers for any convex domain that can be constructed inductively from simpler domains using convexity-preserving operations.
%We discuss this in Section~\ref{???}, and present approaches that are more general and have other advantages over the prior approaches.

\section{Regret Circuits}

In this paper, we introduce \emph{regret circuits}. They are composed of independent regret minimizers connected by wires on which the loss functions and decisions can flow. Regret circuits encode how the inputs and outputs of multiple regret minimizers can be combined to achieve a goal, in a divide-and-conquer fashion, and help simplify the design and analysis of regret-minimizing algorithms. Using the constructions that we will present, one can compose different regret circuits and produce increasingly complex circuits.

The regret circuits approach has several advantages that make it appealing when compared to other, more monolithic, approaches. For one, by treating every regret minimizer that appears in a regret circuit as an independent black box, our approach enables one to select the best individual algorithm for each of them. Second, our framework is amenable to pruning or warm-starting techniques in different parts of the circuit, and substituting one or more parts of the circuit with an approximation. Finally, regret circuits can be easily run in distributed and parallel environments.

\begin{figure}[H]
  \centering\includegraphics[scale=.9]{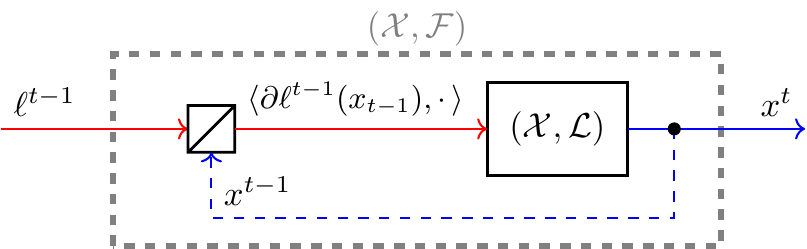}
  \caption{Regret circuit representing the construction of an $(\cX, \cF)$-regret minimizer using an $(\cX, \cL)$-regret minimizer.}
  \label{circ:linear loss}
\end{figure}

We will express regret circuits pictorially through block diagrams.
We will use the following conventions when drawing regret circuits:
\begin{itemize}
    \item an $(\cX, \cF)$-regret minimizer is drawn as a box
    \begin{center}%
    \begin{tikzpicture}[scale=.9]
            \draw[thick] (-.5,0) rectangle (1, 1);
            \node at (.25, .5) {$(\cX,\cF)$};
            \draw[red,semithick,->] (-1.5,.5) --node[above,black]{$\ell^{t-1}$} (-.5,.5);
            \draw[blue,semithick,->] (1,.5) --node[above right,black]{$x^t$} (1.9,.5);
    \end{tikzpicture}%
    \end{center}
    where the input (red) arrow represents the loss at a generic time $t-1$, while the output (blue) arrow represents the decision produced at time $t$;
    \item the symbol \computeloss{} is used to denote an operation that constructs or manipulates one or more loss functions;
    \item the symbol \computerecc{} is used to denote an operation that combines or manipulates one or more decisions;
    \item the symbol \computesum{} denotes an \emph{adder}, that is a node that outputs the sum of all its inputs;
    \item dashed arrows denote decisions that originate from the previous iteration.
\end{itemize}
As an example, consider the construction of Section~\ref{sec:universality linear losses}, where we showed how one can construct a regret minimizer for generic convex loss functions from any regret minimizer for linear loss functions. Figure~\ref{circ:linear loss} shows how that construction can be expressed as a regret circuit.

% \begin{lemma}\label{lem:linear gradient}
%   A $(\cX, \cF)$-regret minimizer can be constructed from a $(\cX, \cL)$-regret minimizer, provided that the functions in $\cF$ have bounded subgradients and $\cX$ has bounded diameter.
% \end{lemma}

%\subsection{Set shifting}
%If we have a $(\cX,\cL)$-regret minimizer $h$, but would like to minimize regret over $(\cY,\cL)$ where $\cY=\cX+b$ for some vector $b$, then we can use $h$ directly: since losses are linear subtracting a constant vector from each point has no effect on regret, and so we simply shift $\cY$ to $\cX$ via the mapping $x=y-b$.
%
%Due to the above we can always assume without loss of generality that $0\in \cX$; if it is not then we can simply subtract any $x\in\cX$ from $\cX$ in order to make it so.

\section{Circuit Construction for Operations that Enlarge or Transform Sets}
\label{sec:calculus 1}
In this section, we begin the construction of our \emph{calculus of regret minimization}. Given regret minimizers for two closed convex sets $\cX$ and $\cY$, we show how to construct a regret minimization for sets obtained via convexity-preserving operations on $\cX$ and $\cY$. In this section, we focus on operations that take one or more sets and produce a regret minimizer for a larger set---this is the case, for instance, of convex hulls, Cartesian products, and Minkowski sums.

As explained in Section~\ref{sec:universality linear losses}, one can extend any $(\cX, \cL)$-regret minimizer to handle more expressive loss functionals.
%In fact, once a $(\cX, \cL)$-regret minimizer is known, extending it to a $(\cX, \cF)$-regret minimizer is mostly a mechanical task.
Therefore, in the rest of the paper, we focus on $(\cX, \cL)$-regret minimizers.
\subsection{Cartesian Product}
\begin{figure}[ht]
  \centering\includegraphics[scale=.9]{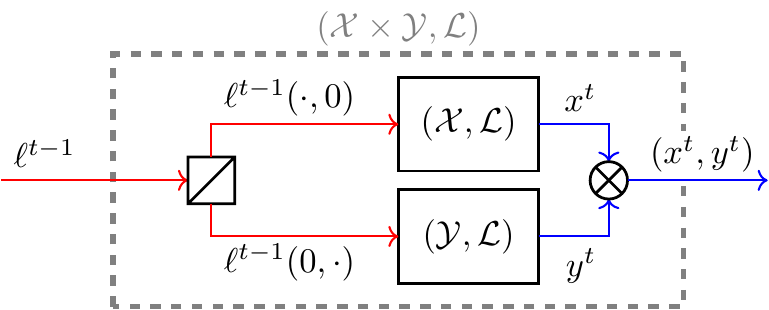}
  \caption{Regret circuit for the Cartesian product $\cX \times \cY$.}
  \label{circ:cartesian product}
\end{figure}

In this section, we show how to combine an $(\cX, \cL)$- and a $(\cY, \cL)$-regret minimizer to form an $(\cX \times \cY, \cL)$-regret minimizer.
%
%As we have already observed, we can assume without loss of generality that $(0,0) \in (\cX, \cY)$. Hence,
Any linear function $\ell: \cX \times \cY \to \bbR $ can be written as
$
  \ell(x, y) = \ell_\cX(x) + \ell_\cY(y)
$
where the linear functions $\ell_\cX : \cX \to\bbR$ and $\ell_\cY:\cY\to\bbR$ are defined as $\ell_\cX : x \mapsto\ell(x, 0)$ and $\ell_\cY: y\mapsto \ell(0, y)$.
It is immediate to verify that
\begin{align*}
  R^T_{(\cX\times\cY,\cL)}
  &= \left(\sum_{t=1}^T \ell_\cX^t(x^t) - \min_{\hat x \in \cX} \left\{\sum_{t=1}^T \ell_\cX^t(\hat x)\right\}\right)\\[-2mm]
                          &\hspace{1cm} +\left(\sum_{t=1}^T \ell_\cY^t(y^t) - \min_{\hat y \in \cY} \left\{\sum_{t=1}^T \ell_\cY^t(\hat y)\right\}\right)\\[-1mm]
                           &= R^T_{(\cX,\cL)} + R^T_{(\cY,\cL)}.
\end{align*}%
\noindent In other words, it is possible to minimize regret on $\cX \times \cY$ by simply minimizing it on $\cX$ and $\cY$ independently and then combining the decisions, as in Figure~\ref{circ:cartesian product}.

\subsection{Affine Transformation and Minkowski Sum}\label{sec:affine map}
Let $T: E \to F$ be an affine map between two Euclidean spaces $E$ and $F$, and let $\cX \subseteq E$ be a convex and compact set. We now show how an $(\cX, \cL)$-regret minimizer can be employed to construct a $(T(\cX), \cL)$-regret minimizer.

Since every $y\in T(\cX)$ can be written as $y=T(x)$ for some $x\in \cX$, the cumulative regret for a $(T(\cX), \cL)$-regret minimizer can be expressed as
\begin{align*}
  R_{(T(\cX),\cL)}^T &= \sum_{t=1}^T (\ell^t \circ T)(x^t) - \min_{\hat x \in \cX} \left\{ \sum_{t=1}^T (\ell^t \circ T)(\hat x) \right\}\!.
\end{align*}%

\begin{figure}[ht]
  \centering\includegraphics[scale=.9]{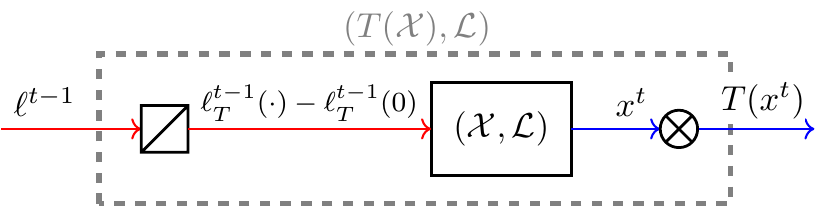}
  \caption{Regret circuit for the image $T(\cX)$ of $\cX$ under the affine transformation $T$.}
  \label{circ:linear trans}
\end{figure}

Since $\ell^t$ and $T$ are affine, their composition $\ell_T^t \defeq \ell^t \circ T$ is also affine. Hence, $R_{(T(\cX),\cL)}^T$ is the same regret as an $(\cX, \cL)$-regret minimizer that observes the linear function $\ell_T^t(\cdot)-\ell_T^t(0)$ instead of $\ell^t$.%
% \footnote{The shifting term $-\ell_T^t(0)$ is needed to ensure that the loss function observed by the $(\cX,\cL)$-regret minimizer is linear rather than affine. This shift does not affect the cumulative regret of the minimizer because the shift occurs both before and after the minus sign in Equation~\ref{eq:regret defn}.}
The construction is summarized by the circuit in Figure~\ref{circ:linear trans}.
%It holds that $\|\ell^t_T\| \le \|\ell^t\|\cdot\|T\|$.

\begin{figure}[ht]
  \centering\includegraphics[scale=.9]{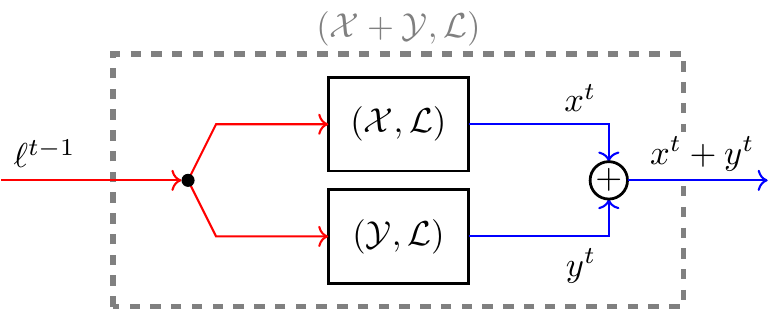}
  \caption{Regret circuit for the Minkowski sum $\cX + \cY$ (Section~\ref{sec:affine map}).}
  \label{circ:minkowski sum}
\end{figure}

As an application, we use the above construction to form a regret minimizer for the Minkowski sum $\cX + \cY := \{x+y: x\in\cX, y\in\cY\}$ of two sets. Indeed, note that
$\cX + \cY = \sigma(\cX \times \cY)$, where $\sigma: (x, y) \mapsto x + y$ is a linear map. Hence, we can combine the construction in this section together with the construction of the Cartesian product (Figure~\ref{circ:cartesian product}). See Figure~\ref{circ:minkowski sum} for the resulting circuit.

\subsection{Convex Hull}
In this section, we show how to combine an $(\cX, \cL)$- and a $(\cY, \cL)$-regret minimizer to form a $(\co\{\cX, \cY\}, \cL)$-regret minimizer, where $\co$ denotes the convex hull operation,
\[
  \co\{\cX, \cY\} = \{\lambda_1 x + \lambda_2 y\ :\ x\in\cX, y\in\cY, (\lambda_1, \lambda_2) \in \Delta^2 \},
\]%
and $\Delta^2$ is the two-dimensional \emph{simplex}
\[
\Delta^2 \defeq \{(\lambda_1, \lambda_2) \in \bbR^2_+: \lambda_1 + \lambda_2 = 1\}.
\]%
Hence, we can think of a $(\co\{\cX, \cY\}, \cL)$-regret minimizer as picking a triple $(\lambda^t, x^t, y^t) \in \Delta^2\times \cX\times \cY$ at each time point $t$. Using the linearity of the loss functions,% the cumulative regret can be written as
\begin{align*}
  R^T_{(\co\{\cX,\cY\},\cL)} &= \left( \sum_{t=1}^T \lambda_1^t\ell^t(x^t) + \lambda_2^t\ell^t(y^t) \right)\\[-3mm]
                             &\hspace{.6cm} -\!\!\!\min_{\substack{\hat \lambda \in \Delta^2\\\hat x \in \cX,\hat y \in \cY}}\!\left\{ {\hat\lambda_1}\sum_{t=1}^T \ell^t(\hat x) + \hat\lambda_2\sum_{t=1}^T \ell^t(\hat y)\right\}\!.
\end{align*}%
Now, we make two crucial observations. First,
{\allowdisplaybreaks\begin{align*}
\min_{\substack{\hat \lambda \in \Delta^2\\\hat x \in \cX,\hat y \in \cY}}\!\left\{ {\hat\lambda_1}\sum_{t=1}^T \ell^t(\hat x) + \hat\lambda_2\sum_{t=1}^T \ell^t(\hat y)\right\}\\[-1mm]
= \min_{\hat \lambda \in \Delta^2}\left\{ {\hat\lambda_1}\min_{\hat x\in\cX}\left\{\sum_{t=1}^T \ell^t(\hat x)\right\} + \hat\lambda_2\min_{y\in\cY}\left\{\sum_{t=1}^T \ell^t(\hat y)\right\}\right\}\!,
\end{align*}}%
since all components of $\hat\lambda$ are non-negative.
Second, the inner minimization problem over $\cX$ is related to the cumulative regret $R_{\cX,\cL)}^T$ of the $(\cX,\cL)$-regret minimizer that observes the loss functions $\ell^t$ as follows:
\begin{align*}
  \min_{\hat x\in\cX}\left\{\sum_{t=1}^T \ell^t(\hat x)\!\right\} &= -R_{(\cX, \cL)}^T + \sum_{t=1}^T \ell^t(x^t).
\end{align*}%
(An analogous relationship holds for $\cY$.) Combining the two observations, we can write
\begin{align*}
  &R^T_{(\co\{\cX,\cY\},\cL)} = \left(\sum_{t=1}^T \lambda_1^t\ell^t(x^t) + \lambda_2^t\ell^t(y^t)\right)\\[-2mm]
  &\hspace{.0cm} -\!\!\min_{\hat \lambda\in \Delta^2}\!\!\left\{\!\!\left(\sum_{t=1}^T {\hat\lambda_1} \ell^t(x^t) \!+\! \hat\lambda_2 \ell^t(y^t)\!\!\right)\!\!-\!\!\left(\!{\hat\lambda_1}R_{(\cX, \cL)}^T \!+\! \hat\lambda_2R_{(\cY, \cL)}^T\!\right)\!\!\right\}\!\!.
\end{align*}%
Using the fact that $\min (f + g) \ge \min f + \min g$, %
% we can break the minimum in the above equation as
% \begin{align*}
%   &-\min_{\hat \lambda \in \Delta^2}\!\left\{\!\left(\sum_{t=1}^T {\hat\lambda_1} \ell^t(x^t) + \hat\lambda_2 \ell^t(y^t)\right)\right.\\ &\hspace{3.5cm}-\left({\hat\lambda_1}R_{(\cX, \cL)}^T + \hat\lambda_2R_{(\cY, \cL)}^T\right)\!\!\Bigg\}\\
%   &\le -\min_{\hat \lambda \in \Delta^2}\!\left\{\!\left(\sum_{t=1}^T {\hat\lambda_1} \ell^t(x^t) +\hat\lambda_2\ell^t(y^t)\right)\right\}\\ &\hspace{4cm} + \max\{R_{(\cX, \cL)}^T, R_{(\cY, \cL)}^T\}.
% \end{align*}%
% Finally, introducing the quantity
%
 and introducing the quantity
\begin{align*}
  R_{(\Delta^2,\cL)}^T &\defeq \left(\sum_{t=1}^T \lambda_1^t\ell^t(x^t) + \lambda_2^t\ell^t(y^t)\right)\\[-2mm]
                    &\hspace{1.3cm}-\!\!\min_{\hat \lambda \in \Delta^2}\!\left\{\!\left(\sum_{t=1}^T {\hat\lambda_1} \ell^t(x^t) +\hat\lambda_2\ell^t(y^t)\right)\!\right\}\!,
\end{align*}%
we conclude that
\begin{equation}\label{eq:ch bound}
 R^T_{(\co\{\cX,\cY\},\cL)} \le R^T_{(\Delta^2,\cL)} + \max\{R_{(\cX, \cL)}^T, R_{(\cY, \cL)}^T\}.
\end{equation}%

\begin{figure}[h]
  \centering\includegraphics[scale=.89]{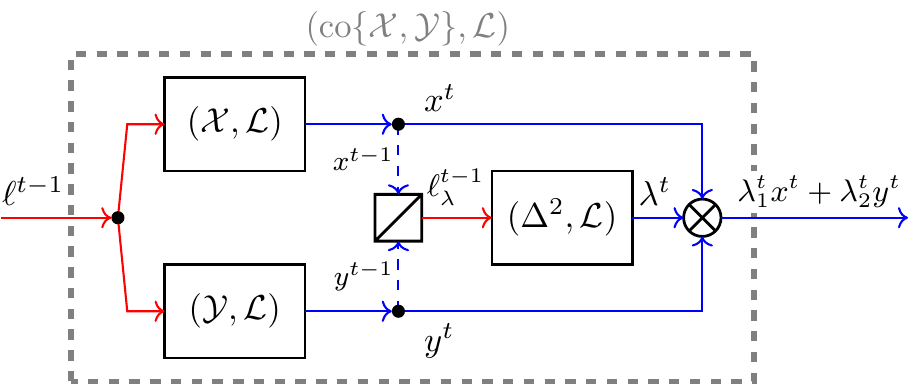}
  \caption{Regret circuit for the convex hull $\co\{\cX, \cY\}$. The loss function $\ell^t_\lambda$ is defined in Equation~\eqref{eq:ell lambda}.}% As usual, dashed arrows represent flows that originate from the previous iteration.}
  \label{circ:convex hull}
\end{figure}

The introduced quantity, $R_{(\Delta^2,\cL)}^T$, is the cumulative regret of a $(\Delta^2,\cL)$-regret minimizer that, at each time instant $t$, observes the (linear) loss function
\begin{equation}\label{eq:ell lambda}
  \ell^t_\lambda : \Delta^2 \ni (\lambda_1,\lambda_2) \mapsto \lambda_1 \ell^t(x^t) + \lambda_2\ell^t(y^t).
\end{equation}%
Intuitively, this means that in order to make ``good decisions'' in the convex hull $\co\{\cX, \cY\}$, we can let two independent $(\cX, \cL)$- and $(\cY, \cL)$-regret minimizers pick good decisions in $\cX$ and $\cY$ respectively, and then use a third regret minimizer that decides how to ``mix'' the two outputs. This way, we break the task of picking the next recommended triple $(\lambda^t, x^t, y^t)$ into three different subproblems, two of which can be run independently.
Equation~\eqref{eq:ch bound} guarantees that if all three regrets $\{R_{(\Delta^2,\cL)}^T, R_{(\cX, \cL)}^T, R_{(\cY, \cL)}^T\}$ grow sublinearly, then so does $R_{(\co\{\cX, \cY\},\cL)}^T$.
Figure~\ref{circ:convex hull} shows the regret circuit that corresponds to our construction above.

\textbf{Extending to multiple set. }
% Since the convex hull operation is associative, we can handle the convex hull $\co\{\cX_1,\dots,\cX_n\}$ of a finite number of sets by iteratively adding one set at a time. Equivalently, o
The construction shown in Figure~\ref{circ:convex hull} can be extended to handle the convex hull $\co\{\cX_1,\dots,\cX_n\}$ of $n$ sets as follows. First, the input loss function $\ell^{t-1}$ is fed into all the $(\cX_i, \cL)$-regret minimizers ($i=1,\dots, n$). Then, the loss function $\ell_\lambda^t$, defined as
\[
  \ell_\lambda^{t} : \Delta^n \ni (\lambda_1, \dots, \lambda_n) \mapsto \lambda_1 x_1^{t} + \dots + \lambda_n x_n^{t},
\]%
is input into a $(\Delta^n,\cL)$-regret minimizer, where $\Delta^n$ is the $n$-dimensional simplex.
Finally, at each time instant $t$, the $n$ decisions $x_1^t, \dots, x_n^t$ output by the $(\cX_i, \cL)$-regret minimizers are combined with the decision $\lambda^t$ output by the $(\Delta^n,\cL)$-regret minimizer to form $\lambda^t_1 x_1^t + \dots + \lambda_n^t x_n^t$.

\textbf{$V$-polytopes. }
Our construction can be directly applied to construct an $(\cX, \cL)$-regret minimizer for a $V$-polytope $\cX = \co\{v_1, \dots, v_n\}$ where $v_1, \dots, v_n$ are $n$ points in a Euclidean space $E$. Of course, any $(\{v_i\}, \cL)$-regret minimizer outputs the constant decision $v_i$. Hence, our construction (Figure~\ref{circ:convex hull}) reduces to a single $(\Delta^n, \cL)$-regret minimizer that observes the (linear) loss function
\[
  \ell_\lambda^t : \Delta^n \ni (\lambda_1, \dots, \lambda_n) \mapsto \lambda_1 \ell^t(v_1) + \dots + \lambda_n \ell^t(v_n).
\]
The observation that a regret minimizer over a simplex can be used to minimize regret over a $V$-polytope already appeared in~\citet{Zinkevich03:Online} and~\citet[Theorem~3]{Farina17:Regret}. 

\section{Application: Derivation of CFR}

We now show that these constructions can be used to construct the CFR framework.
The first thing to note is that the strategy space of a single player in an EFG
is a treeplex, which can be viewed recursively as a series of convex hull and
Cartesian product operations. This perspective is also used when
  constructing distance functions for first-order methods for
  EFGs~\citep{Hoda10:Smoothing,Kroer15:Faster,Kroer18:Faster}.
In particular,
an information set is viewed as an $n$-dimensional convex hull (since the sum of
probabilities over actions is $1$), where each action $a$ at the information set
corresponds to a treeplex $\cX_a$ representing the set of possible information
sets coming after $a$ (in order to perform the convex hull operation, we create a
new, larger representation of $\cX_a$ so that the dimension is the same for all
$a$, described in detail below). The Cartesian product operation is used to
represent multiple potential information sets being arrived at (for example
different hands dealt in a poker game).

\begin{figure}[t]
  \centering\includegraphics[width=.95\linewidth]{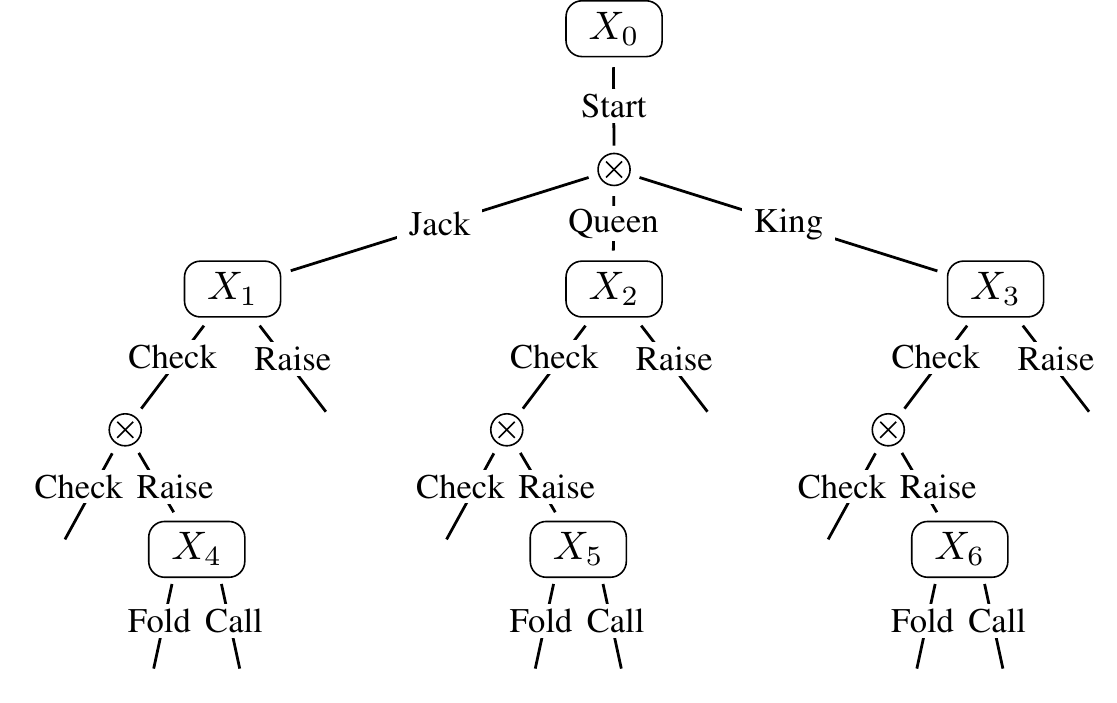}
  \vspace{-3mm}
  \caption{Treeplex for the first player in the game of Kuhn poker. Each $X_i$ represents a convex hull over the treeplexes below, while \treeplexproduct{} denotes the Cartesian product operation.}
  \label{fig:kuhn treeplex player1}
\end{figure}

Figure~\ref{fig:kuhn treeplex player1} shows an example. Each information set $X_i$ (except
$X_0$) corresponds to a $2$-dimensional convex hull over two treeplexes, one of
which is always empty (that is, a leaf node). Each \treeplexproduct{} is a Cartesian
product. The top-most \treeplexproduct{} represents the three possible hands that the
player may have when making their first decision. The second layer of Cartesian
products represent actions taken by the opponent.

\begin{figure}[t]
    \centering\includegraphics[scale=.92]{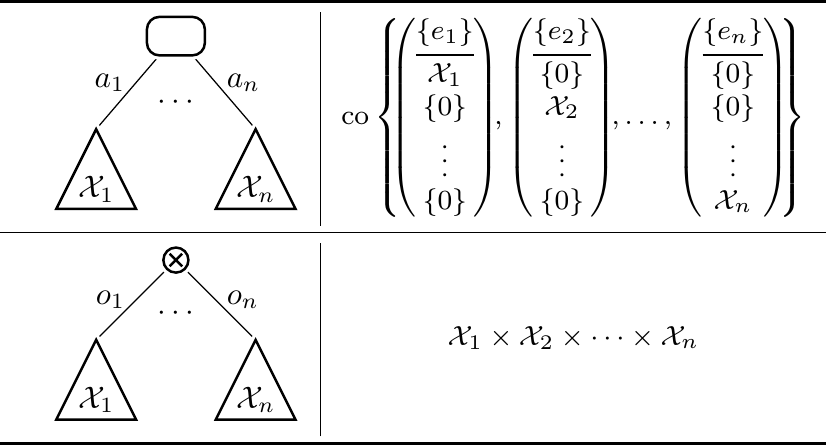}
    \caption{Inductive treeplex construction rules. $e_i \in \bbR^n$ contains a 1 at index $i$, and 0 everywhere else.}
    \label{fig:treeplex math}
\end{figure}

The information-set construction is as follows: let $I$ be the information set under construction, and $A_I$ the set of actions. Each action $a\in A_I$ has some, potentially empty, treeplex $\cX_a$ beneath it; let $n_a$ be the dimension of that treeplex. We cannot form a convex hull over $\{\cX_a\}_{a\in A_I}$ directly since the sets are not of the same dimension, and we do not wish to average across different strategy spaces. Instead, we create a new convex set $\cX_a'\in \R^{|A_I|+\sum_{a\in A_I}n_a}$ for each $a$. The first $|A_I|$ indices correspond to the actions in $A_I$, and each $\cX_a$ gets its own subset of indices. For each $x\in \cX_a$ there is a corresponding $x'\in \cX_a'$; $x'$ has a $1$ at the index of $a$, $x$ at the indices corresponding to $\cX_a$, and $0$ everywhere else. The convex hull is constructed over the set $\{\cX_a'\}_a$, which gives exactly the treeplex rooted at $I$. The Cartesian product is easy and can be done over a given set of treeplexes rooted at information sets $I_1,\ldots,I_n$.
The inductive construction rules for the treeplex are given in Figure~\ref{fig:treeplex math}.
In fact, one can prove that the $\ell_\lambda$ loss functions defined in Equation~\eqref{eq:ell lambda} are exactly the counterfactual loss functions defined in the original CFR paper~\cite{Zinkevich07:Regret}.
If we use as our loss function the gradient $Ay^t$ where $y^t$ is the opponent's strategy at iteration $t$, and then apply our expressions for the Cartesian-product and convex-hull regrets inductively, it follows from \eqref{eq:ell lambda} that the loss function associated with each action is exactly the negative counterfactual value. Finally, the average treeplex strategy as per Theorem~\ref{thm:folk theorem} coincides with the per-information-set averaging used in standard CFR expositions (e.g.,~\citep{Zinkevich07:Regret}).


% As an example, we describe how we would construct the convex hull for $X_2$ in
% Figure~\ref{fig:kuhn treeplex player1}. There are two actions \{check, raise\}.
% The raise treeplex $\cX_{\text{raise}}$ is empty, and so $n_{\text{raise}}=0$.
% The check treeplex has one information set and $n_{\text{check}}=2$. The space
% we will perform the convex hull in is $\R^{4}$, and we set the index of
% check and raise as 1 and 2 respectively, and put the $\cX_{\text{check}}$ treeplex at indices 3 and 4
% ($\cX_{\text{raise}}$ is empty and so requires no indices).
% $\cX_{\text{raise}}'$ is the set $\{[0,1,0,0]\}$, while $\cX_{\text{check}}'$ is
% the set $\{[1,0,x_1,x_2]: x \in \Delta^2\}$, where $\Delta^2$ represents all probability distributions over fold/call at $X_5$.

% To formally prove the construction, one uses the fact that there exists a bottom-up ordering of the treeplex. The treeplex can then be constructed recursively by applying the Cartesian product and convex hull operations as described above.

%%% Local Variables:
%%% mode: latex
%%% TeX-master: "../main"
%%% End:

\section{Circuit Construction for Operations that Constrain Sets}
Unlike Section~\ref{sec:calculus 1}, in this section we deal with operations that curtail the set of decisions that can be output by our regret minimizer.
Section~\ref{sec:lagrangian relaxation} and~\ref{sec:intersection} propose two different constructions, and Section~\ref{sec:comparison of intersection construction} discusses the merits and drawbacks of the two.
\subsection{Constraint Enforcement via Lagrangian Relaxation}
\label{sec:lagrangian relaxation}
\begin{figure}[ht]
  \centering\includegraphics[scale=.9]{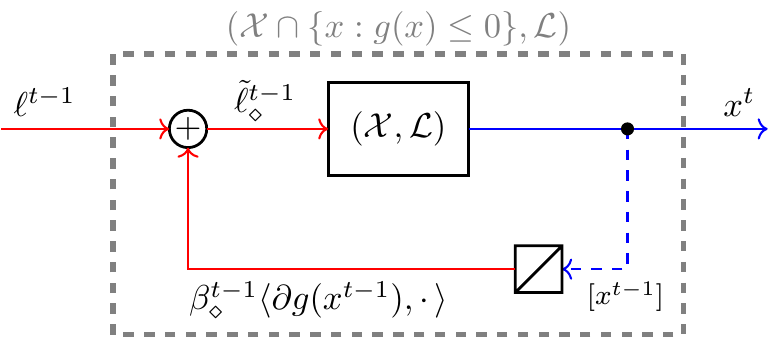}
  \caption{Regret circuit for the Lagrangian relaxation construction (Section~\ref{sec:lagrangian relaxation}) for the constrained set $\cX\cap\{x: g(x)\le 0\}$.}
  \label{circ:lagrangian relaxation}
\end{figure}

Suppose that we want to construct an $(\cX \cap \{x: g(x) \!\le\! 0\}, \cL)$-regret minimizer, where $g$ is a convex function, but we only possess an $(\cX,\cL)$-regret minimizer. %In order to make the problem interesting, we assume that $g(x) > 0$ for at least one $x \in \cX$.
One natural idea is to use the latter to approximate the former, by penalizing any choice of $x\in\cX$ such that $g(x) > 0$. In particular, it seems natural to introduce the penalized loss function
\[
  \tilde\ell^t: \cX \ni x \mapsto \ell^t(x) + \beta^t \max\{ 0, g(x)\},
\]%
where $\beta^t$ is a (large) positive constant that can change over time. This approach is reminiscent of Lagrangian relaxation.
The loss function $\tilde\ell^t$ is not linear, and as such it cannot be handled as is by our $(\cX, \cL)$-regret minimizer. However, as we have observed in Section~\ref{sec:universality linear losses}, the regret induced by $\tilde\ell^t$ can be minimized by our $(\cX,\cL)$-regret minimizer if that observes the ``linearized'' loss function
\begin{align*}
  & \tilde\ell_\diamond^t : \cX \ni x \mapsto \ell^t(x) + \beta_\diamond^t \langle \partial g(x^t), x \rangle,\qquad\\[-1mm]
  \text{where}\quad &
  \beta_\diamond^t \defeq \begin{cases}
    \beta^t & \text{if } g(x^t) > 0\\[-1mm]
    0 & \text{otherwise}.
  \end{cases}
\end{align*}
Figure~\ref{circ:lagrangian relaxation} shows the regret circuit corresponding to the construction described so far.

In the rest of this subsection we analyze in what sense small cumulative regret implies that the constraint $g(x) \le 0$ is satisfied.
Let $R_{(\cX,\cL)}^T$ be the cumulative regret of our $(\cX, \cL)$-regret minimizer. Introducing $\cX_g \defeq \cX \cap\{x: g(x) \le 0\}$ and $\tau_g \defeq \{t\in\{1,\dots, T\}: g(x^t)\!>\!0\}$,
\begin{align}
  R_{(\cX,\cL)}^T &= \sum_{t=1}^T \tilde\ell_\diamond^t(x^t) - \min_{\hat x\in\cX}\left\{\sum_{t=1}^T \tilde\ell_\diamond^t(\hat x)\right\}\nonumber\\[-1mm]
       &\ge \sum_{t=1}^T \ell^t(x^t) + \sum_{t\in\tau_g} \beta^t g(x^t)\nonumber\\[-2mm]
       &\hspace{0.92cm}-\min_{\hat x \in \cX} \left\{\sum_{t=1}^T \ell^t(\hat x) \!+\! \left(\sum_{t=1}^T \beta_i\!\right)\!\max\{0, g(\hat x)\}\!\right\}\nonumber\\[-1mm]
       &\ge \left(\sum_{t=1}^T\ell^t(x^t)\!-\!\min_{\hat x\in \cX_g} \sum_{t=1}^T \ell^t(\hat x)\!\right) \!+\!\!  \sum_{t\in\tau_g} \beta^t g(x^t),\label{eq:lagrangian relaxation regret}
\end{align}%
where the first inequality is by \eqref{eq:linearization bound} and the second inequality comes from simply restricting the domain of the minimization from $\cX$ to $\cX_g$.\footnote{It may tempting to recognize in the term in parentheses in~\eqref{eq:lagrangian relaxation regret} the cumulative regret of an\! $(\cX_g,\! \cL)$-regret minimizer.\! This would be incorrect: the decisions $x^t$ are \emph{not} guaranteed to satisfy $g(x^t) \le 0$.} Thus, if the $\beta^t$ are sufficiently large, the average decision $\bar x \defeq \frac{1}{B}(\beta^1 x^1 + \dots + \beta^T x^T)$ where $B \defeq \beta^1 + \dots + \beta^T$ satisfies
\begin{align*}
  \max\{0, g\}(\bar x) &\le \frac{1}{B}\sum_{t=1}^T \beta^t \max\{0, g\}(x^t)= \frac{1}{B}\sum_{t\in\tau_g} \beta^t g(x^t)\\[-2mm]
            &\le \frac{1}{B}\!\left(\!R_{(\cX,\cL)}^T + \min_{\hat x\in \cX_g} \left\{ \sum_{t=1}^T \ell^t(\hat x - x^t) \!\right\}\!\right)\!,
\end{align*}%
where the first inequality follows by convexity of $\max\{0, g\}$, and the second inequality follows by \eqref{eq:lagrangian relaxation regret}.

If $\sum_{t=1}^T \beta_i \gg T L D$, where $L$ is an upper bound on the norm of the loss functions $\ell^1, \dots, \ell^T$ and $D$ is an upper bound on the diameter of $\cX$, then $\max\{0, g(\bar x)\} \to 0$ as $T \to \infty$, that is, the constraint is satisfied at least by the average in the limit.
If $L$ and $D$ are known ahead of time, one practical way to guarantee $\sum_{t=1}^T \beta_i \gg T L D$ is to choose $\beta^t = \kappa LD$ where $\kappa>0$ is a large constant. This guarantees that in the limit, small cumulative regret implies that the average strategy approximately satisfies the constraint and satisfies Hannan consistency.
Formally:
\begin{theorem}
The decisions $\{x^t\}$ produced by a regret minimizer that observes loss functions $\{{\tilde\ell}^t_\diamond\}$ where $\beta^t = \kappa LD$ satisfy the following two properties:
\begin{itemize}
  \item Approximate feasibility: $\displaystyle g\!\left(\frac{1}{T}\sum_{t=1}^T x^t\right) \le \frac{1}{\kappa} + o(1).$
  \item Hannan consistency with respect to \{$\ell^t$\}: \[\sum_{t=1}^T\ell^t(x^t)\!-\!\min_{\hat x\in \cX_g} \left\{\sum_{t=1}^T \ell^t(\hat x) \right\} = o(T).\]
\end{itemize}
\end{theorem}

Alternatively, the $\beta^t$ can be chosen by a regret minimizer which sees the constraint violation $\max\{0, g(x^t)\}$ at time $t$ as its loss function.

\subsection{Intersection with a Closed Convex Set}
\label{sec:intersection}
In this subsection we consider constructing an $(\cX \cap \cY, \cL)$-regret minimizer from an $(\cX, \cL)$-regret minimizer, where $\cY$ is a closed convex set such that $\cX \cap \cY \ne \emptyset$.
As it turns out, this is always possible, and can be done by letting the $(\cX, \cL)$-regret minimizer give decisions in $\cX$, and then \emph{projecting} them onto the intersection $\cX\cap\cY$.

\begin{figure}[ht]
  \centering\includegraphics[scale=.9]{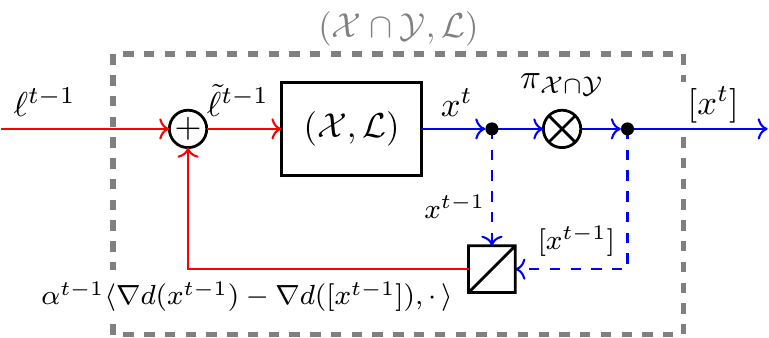}
  \caption{Regret circuit representing the construction of an $(\cX\cap\cY, \cL)$-regret minimizer using a $(\cX, \cL)$-regret minimizer.}
  \label{circ:intersection}
\end{figure}

We will use a \emph{Bregman divergence}
$\displaystyle
  D(y \| x) \defeq d(y) - d(x) - \langle \nabla d(x), y - x\rangle
$
as our notion of (generalized) distance between the points $x$ and $y$, where the \emph{distance generating function} (DGF) $d$ is $\mu$-strongly convex and $\beta$-smooth (that is, $d$ is differentiable and its gradient is Lipschitz continuous with Lipschitz constant $\beta$). Our construction makes no further assumptions on $d$, so the most appropriate DGF can be used for the application at hand. When $d(x) = \|x\|_2^2$ we obtain $D(x \| y) = \|x - y\|_2^2$, so we recover the usual Euclidean distance between $x$ and $y$. In accordance with our generalized notion of distance, we define the projection of a point $x \in \cX$ onto $\cX \cap \cY$ as
$\displaystyle
  \pi_{\cX \cap \cY}(x) = \argmin_{y\in \cX \cap \cY} D(y \| x).
$
For ease of notation, we will denote the projection of $x$ onto $\cX\cap\cY$ as $[x]$; since $\cX\cap\cY$ is closed and convex, and since $D(\cdot \| x)$ is strongly convex, such projection exists and is unique. As usual, the cumulative regret of the $(\cX\cap\cY,\cL)$-minimizer is
\begin{align}
  R_{(\cX\cap\cY, \cL)}^T &\!= \sum_{t=1}^T \ell^t([x^t]) - \min_{\hat x \in \cX\cap\cY}\left\{\sum_{t=1}^T \ell^t(\hat x) \right\}\nonumber\\[-1mm]
  &\!= \sum_{t=1}^T \ell^t([x^t] \!-\! x^t) -\!\! \min_{\hat x \in \cX\cap\cY}\!\left\{\sum_{t=1}^T \ell^t(\hat x\!-\!x^t)\!\right\}\!\!,\label{eq:alpha regret}
\end{align}%
where the second equality holds by linearity of $\ell^t$. The first-order optimality condition for the projection problem is
\[
  \langle \nabla d(x^t) - \nabla d([x^t]), \hat x - [x^t]\rangle \le 0 \quad\forall\, \hat x \in \cX \cap \cY.
\]%
Consequently, provided $\alpha^t \ge 0$ for all $t$,
\begin{align}
  &\min_{\hat x \in \cX \cap \cY} \left\{ \sum_{t=1}^T \ell^t(\hat x - x^t)\right\} \ge \min_{\hat x \in \cX} \left\{\sum_{t=1}^T\ell^t(\hat x - x^t)\right.\nonumber\\[-2mm]
  &\hspace{1.6cm} + \sum_{t=1}^T \alpha^t \langle \nabla d(x^t) - \nabla d([x^t]), \hat x - [x^t]\rangle\!\Bigg\}.\label{eq:alpha min bound}
\end{align}%
The role of the $\alpha^t$ coefficients is to penalize choices of $x^t$ that are in $\cX \setminus \cY$. In particular, if
\begin{equation}\label{eq:alpha t condition}
  \frac{1}{\mu}\sum_{t=1}^T \ell_t([x^t] - x^t)
  \le
  \sum_{t=1}^T \alpha^t \|[x^t] - x^t\|^2,
\end{equation}%
then, by $\mu$-strong convexity of $d$, we have
\begin{equation}\label{eq:alpha t condition 2}
  \sum_{t=1}^T \ell_t([x^t] \!-\! x^t)
  \le
  \sum_{t=1}^T \alpha^t \langle \nabla d(x^t) \!-\! \nabla d([x^t]), x^t \!-\! [x^t] \rangle.
\end{equation}
Substituting~\eqref{eq:alpha t condition 2} and~\eqref{eq:alpha min bound} into Equation~\eqref{eq:alpha regret} we get
{\allowdisplaybreaks\begin{align*}
    &R_{(\cX \cap \cY,\cL)}^T
    %&\le \sum_{t=1}^T \alpha^t \|[x^t] - x^t\|^2 - \min_{\hat x\in \cX \cap \cY} \left\{ \sum_{t=1}^T \ell^t(\hat x - x^t)\right.\\[-2mm]
    %&\hspace{2.62cm} + \sum_{t=1}^T  \alpha^t \langle x^t - [x^t], \hat x - [x^t]\rangle\Bigg\}\\[-2mm]
    % &\le \left(\sum_{t=1}^T \ell^t(x^t) + \alpha^t\langle x^t - [x^t], x^t \rangle \right)\\[-1mm]
    % &\hspace{1.4cm} -\min_{\hat x\in \cX \cap \cY} \left\{ \sum_{t=1}^T \ell^t(\hat x) + \alpha^t\langle x^t - [x^t], \hat x\rangle \right\}\!\\
    \le \left(\sum_{t=1}^T \ell^t(x^t) + \alpha^t\langle \nabla d(x^t) - \nabla d([x^t]), x^t \rangle \right)\\
    &\hspace{1.4cm} -\min_{\hat x\in \cX} \left\{ \sum_{t=1}^T \ell^t(\hat x) + \alpha^t\langle \nabla d(x^t) - \nabla d([x^t]), \hat x\rangle\! \right\}\!\!,
\end{align*}}%
which is the regret observed by an $(\cX, \cL)$-regret minimizer that at each time $t$ observes the linear loss function
\begin{equation}\label{eq:tilde ell intersection}
  {\tilde\ell}^t : x \mapsto \ell^t(x) + \alpha^t \langle \nabla d(x^t) - \nabla d([x^t]), x\rangle.
\end{equation}%
Hence, as long as condition~\eqref{eq:alpha t condition} holds, the regret circuit of Figure~\ref{circ:intersection} is guaranteed to be Hannan consistent.

On the other hand, condition~\eqref{eq:alpha t condition} can be trivially satisfied by the deterministic choice 
\[
 \alpha^t = \begin{cases}
   0 & \text{if } x^t \in \cX \cap \cY\\[2mm]
   \max\left\{0, \displaystyle\frac{\ell^t([x^t] - x^t)}{\mu\|[x^t] - x^t\|^2}\right\} &\text{otherwise}.
 \end{cases}
\]%
The fact that $\alpha_t$ can be arbitrarily large (when $x^t$ and $[x^t]$ are very close) is not an issue. Indeed, $\alpha^t$ is only used in $\tilde\ell^t$ (Equation~\ref{eq:tilde ell intersection}) and is always multiplied by a term whose magnitude grows proportionally with the distance between $x^t$ and $[x^t]$. In fact, the norm  of the functional $\tilde\ell^t$ is bounded:
\begin{align*}
  \|\tilde\ell^t\| &\le \|\ell^t\| + \left\vert\frac{\ell^t([x^t] - x^t)}{\mu\|[x^t] - x^t\|^2}\right\vert \cdot \|\nabla d(x^t) - \nabla d([x^t])\| \\
                   &\le \|\ell^t\| + \beta\left\vert\frac{\ell^t([x^t] - x^t)}{\mu\|[x^t] - x^t\|^2}\right\vert\cdot \|[x^t] - x^t\| \\
                   %&\le \|\ell^t\| + \beta\left\vert\frac{\ell^t([x^t] - x^t)}{\mu\|[x^t] - x^t\|}\right\vert \\
                   &\le \frac{\beta(1+\mu)}{\mu}\|\ell^t\|,
\end{align*}%
where the second inequality follows by $\beta$-smoothness of $d$.
In other words, our construction dilates the loss functions by at most a factor $\beta(1+\mu)/\mu$.

\subsection{Comparison of the Two Constructions}\label{sec:comparison of intersection construction}
As we pointed out, the decisions in the construction using Lagrangian relaxation only converge to the constrained domain $\cX_g$ \emph{on average}. Thus, formally the construction does \emph{not} provide an $(\cX_g, \cL)$-regret minimizer, but only an approximate one. Section~\ref{sec:intersection} solves this problem by providing a generic construction for an $(\cX \cap \cY,\cL)$-regret minimizer, a strictly more general task. The price to pay is the need for (generalized) projections, a potentially expensive operation.
Thus, the choice of which construction to use reduces to a tradeoff between the computational cost of projecting and the need to have exact versus approximate feasibility with respect to $g$. The right choice depends on the application at hand.
Finally, the construction based on Lagrangian relaxation requires large penalization factors $\beta^t$ in order to work properly. Therefore, the norm of $\tilde\ell_\diamond^t$ can be large, which can complicate the task of minimizing the regret $R_{(\cX, \cL)}^T$. 

\section{Application:\! Handling Strategy Constraints}

When solving EFGs, there may be a need to add additional constraints beyond simply computing feasible strategies:
%Such needs can arise for several reasons:

\begin{itemize}[nolistsep]
\item Opponent modeling. Upon observing repeated play from an opponent, we may wish to constrain our model of their strategy space to reflect such observations. Since observations can %in general
    be consistent with several information sets belonging to the opponent, this requires adding constraints that span across information sets.
\item Bounding probabilities. For example, in a patrolling game we may wish to ensure that a patrol returns to its base at the end of the game with high probability.
\item Nash equilibrium refinement computation. Refinements can be computed, or approximated, via perturbation of the strategy space of each player. For \emph{extensive-form perfect equilibrium} this can be done by lower-bounding the probability of each action at each information set~\citep{Farina17:Extensive}, which can be handled with small modifications to standard CFR or first-order methods~\citep{Farina17:Regret,Kroer17:Smoothing}. However, \emph{quasi-perfect equilibrium} requires perturbations on the probability of sequences of action~\citep{Miltersen10:Computing}, which requires strategy constraints that cross information sets.
\end{itemize}

All the applications above potentially require adding strategy space constraints that span across multiple information sets. Such constraints break the recursive nature of the treeplex, and are thus not easily incorporated into standard regret-minimization or first-order methods for EFG solving.
\citet{Davis19:Solving} propose a Lagrangian relaxation approach called
\emph{Constrained CFR} (CCFR): each strategy constraint is added to the
objective with a Lagrangian multiplier, and a regret minimizer is used to
penalize violation of the strategy
constraints. They prove that if the
regret minimizer for the Lagrange multipliers has the optimal Lagrangian
multipliers as part of their strategy space, the average output strategy converges to an approximate solution to the constrained game. They also prove a bound on the approximate feasibility of the average output strategy when their algorithm is instantiated with Regret Matching~\cite{Hart00:Simple} as the local regret minimizer at each information set.

At least two alternative variants of CFR for EFGs with strategy constraints can be obtained using our framework. First,
%We already proved that CFR can be constructed from
%our framework. In order to additionally support strategy constraints
we can
apply our method for Lagrangian relaxation of $\cX$ and a constraint
$g(x)\leq 0$. Our Lagrangian approach yields as a special case the CCFR algorithm. Our approach supports regret
minimization for the Lagrangian multipliers, as was done in CCFR, since we put no constraints on the form of the $\beta^t$ multipliers. However, our approach is more general in that it also allows instantiation with a fixed choice of multipliers, thus obviating the need for regret minimization.
The second alternative is to apply our construction for the intersection of convex sets (Section~\ref{sec:intersection}), which uses
(generalized) projection onto $\cX \cap \{x: g(x) \leq 0 \}$.
This leads to a different
regret-minimization approach, which has the major advantage that all iterates are feasible, whereas Lagrangian approaches only achieve approximate feasibility. The cost of projection may be nontrivial, and so in general the choice of method depends on the application at hand.

%%% Local Variables:
%%% mode: latex
%%% TeX-master: "../main"
%%% End:

\section{Conclusion and Future Research}

We developed a calculus of regret minimization, which enables the construction
of regret minimizers for composite convex sets that can be inductively expressed as a series
of convexity-preserving operations on simpler sets. We showed that our calculus
can be used to construct the CFR algorithm directly, as well as several of its variants
for the more general case where we have strategy constraints. Our
regret calculus is much more broadly applicable than just EFGs:
%and sequential action spaces:
it applies to any setting where the decision space can be
expressed via the convexity-preserving operations that we support. In the future
we plan to investigate novel applications of our regret calculus. One potential
application would be online portfolio selection with additional constraints
(e.g., exposure constraints across industries); our framework makes it easy to construct such a regret minimizer from any standard online-portfolio-selection algorithm.

%We also plan to use our construction for CFR with strategy constraints to test whether the fact that we have explicit bounds on the Lagrangian multipliers leads to a practically superior algorithm compared to \citet{Davis19:Solving}. We also plan to investigate how much better the projection-based algorithm does in terms of iteration complexity, and whether that makes up for the computational cost of projection.

The approach presented in this paper has a large number of potential future applications. For one,
it would be interesting to apply our approaches of including additional constraints to the computation of quasi-perfect equilibria. Currently the only solver that is fairly scalable is based on an exact, monolithic, custom approach that uses heavy-weight operations such as matrix inversion, etc.~\citep{Farina18:Practical}. Our regret-minimization approach would be the first of its kind for equilibrium refinement that requires constraints that cut across information sets. It obviates the need for the heavy-weight operations, and would still converge to a feasible solution that satisfies an approximate notion of quasi-perfect equilibrium. It would be interesting to study this tradeoff between speed and solution quality.

%%% Local Variables:
%%% mode: latex
%%% TeX-master: "../main"
%%% End:

%\clearpage
\bibliography{dairefs}
\bibliographystyle{custom_arxiv}

\end{document}